\documentclass[letterpaper, 10 pt, conference]{ieeeconf}  

\IEEEoverridecommandlockouts                              

\usepackage{graphics} 
\usepackage{epsfig} 
\usepackage{mathptmx} 
\usepackage{times} 
\usepackage{amsmath} 
\usepackage{amssymb}  

\usepackage{multirow}
\usepackage{makecell}
\usepackage{pifont}
\usepackage{color}
\usepackage{colortbl}
\usepackage{xcolor}
\usepackage[export]{adjustbox}
\usepackage{booktabs}

\usepackage{arydshln}
\usepackage{soul}

\usepackage{glossaries}
\newacronym{fps}{FPS}{Farthest-Point-Sampling}
\newacronym{rs}{RS}{Random-Sampling}
\newacronym{ft3d}{$\mathrm{FT3D_s}$}{FlyingThings3D Subset}

\newacronym{ft3d_bold}{$\mathrm{\mathbf{FT3D_s}}$}{FlyingThings3D Subset_}

\newacronym{lfa}{LFA}{Local-Feature-Aggregation}
\newacronym{knn}{KNN}{K-Nearest-Neighbor}
\newacronym{us}{US}{Up-Sampling}
\newacronym{ds}{DS}{Down-Sampling}
\newacronym{fe}{FE}{Flow-Embedding}
\newacronym{wl}{WL}{Warping-Layer}
\newacronym{max}{*}{Max-Pooling}

\newcommand{\mytilde}{{\raise.17ex\hbox{$\scriptstyle\sim$}}}

\newcommand{\PTDP}{\mbox{\textit{Patch-to-Dilated-Patch}}}
\newcommand\Tstrut{\rule{0pt}{1.6ex}}         
\newcommand\Bstrut{\rule[-0.7ex]{0pt}{0pt}}   
\newcommand{\cmark}{\ding{51}}%
\newcommand{\xmark}{\ding{55}}%

\usepackage{xspace}

\newcommand*{\ie}{i.e.\@\xspace}
\newcommand*{\etal}{et al.\@\xspace}
\newcommand*{\cf}{c.f.\@\xspace}

\graphicspath{{figures/}}

\newcommand{\name}{\mbox{RMS-FlowNet}} 

\usepackage{hyperref}

\usepackage[sort&compress,capitalize,nameinlink]{cleveref}
\hypersetup{
	citecolor=green,
	linkcolor=blue,
	filecolor=magenta,      
	urlcolor=cyan,
}

\title{\LARGE \bf
\name{}: Efficient and Robust Multi-Scale Scene Flow Estimation\\for Large-Scale Point Clouds
}

\author{Ramy Battrawy$^{1}$, Ren{\'e} Schuster$^{1}$, Mohammad-Ali Nikouei Mahani$^{2}$ and Didier Stricker$^{1}$
	\thanks{$^{1}$DFKI -- German Research Center for Artificial Intelligence, Germany:
		{\tt\small firstname.lastname@dfki.de}}%
	\thanks{$^{2}$BMW Group, Germany: \newline 
		{\tt\small Mohammad-Ali.Nikouei-Mahani@bmw.de}}%
}

\begin{document}

\maketitle
\thispagestyle{empty}
\pagestyle{empty}

\begin{abstract} \label{abstract}
    The proposed \name{} is a novel end-to-end learning-based architecture for accurate and efficient scene flow estimation which can operate on point clouds of high density.
    For hierarchical scene flow estimation, the existing methods depend on either expensive \gls{fps} or structure-based scaling which decrease their ability to handle a large number of points.
    Unlike these methods, we base our fully supervised architecture on \gls{rs} for multi-scale scene flow prediction.
	To this end, we propose a novel flow embedding design which can predict more robust scene flow in conjunction with \gls{rs}.
	Exhibiting high accuracy, our \name{} provides a faster prediction than state-of-the-art methods and works efficiently on consecutive dense point clouds of more than 250K points at once.
	Our comprehensive experiments verify the accuracy of \name{} on the established FlyingThings3D data set with different point cloud densities and validate our design choices.
	Additionally, we show that our model presents a competitive ability to generalize towards the real-world scenes of KITTI data set without fine-tuning.
\end{abstract}

\section{Introduction} \label{introduction} \glsresetall
Scene flow estimation is a key computer vision task for the purposes of navigation, planning tasks and autonomous driving systems.   
It concerns itself with the estimation of a 3D motion field with respect to the observer, thereby providing a representation of the dynamic change in the surroundings. 

Most of the popular scene flow methods use monocular images \cite{brickwedde2019mono, hur2020self} or stereo images to couple the geometry reconstruction with scene flow estimation \cite{aleotti2020learning, ilg2018occlusions, jiang2019sense, saxena2019pwoc, schuster2021deep, schuster2017sceneflowfields, yang2020upgrading}.
However, the accuracy of such image-based solutions is still constrained by the images quality and the illumination conditions.

In contrary, LiDAR sensors provide accurate measurements of the geometry (as 3D point clouds) with ongoing developments towards increasing their density (\ie the sensor resolution).
Leveraging this potential is becoming increasingly important for the accurate computation of scene flow from point clouds.

To this end, many existing approaches \cite{liu2019flownet3d, wu2020pointpwc, wang2021hierarchical, kittenplon2021flowstep3d} focus on the 3D domain and present highly accurate scene flow with better generalization compared to the image-based modalities.
Such approaches use \gls{fps} \cite{li2018pointcnn, qi2017pointnet++, wu2019pointconv, zhao2019pointweb} leading to a robust feature extraction and an accurate computation for feature similarities. 
However, the expensive computation of \gls{fps} decreases their capabilities to operate efficiently on dense point clouds.

\begin{figure}[t]
	\begin{center}
		\includegraphics[width=1.0\linewidth]{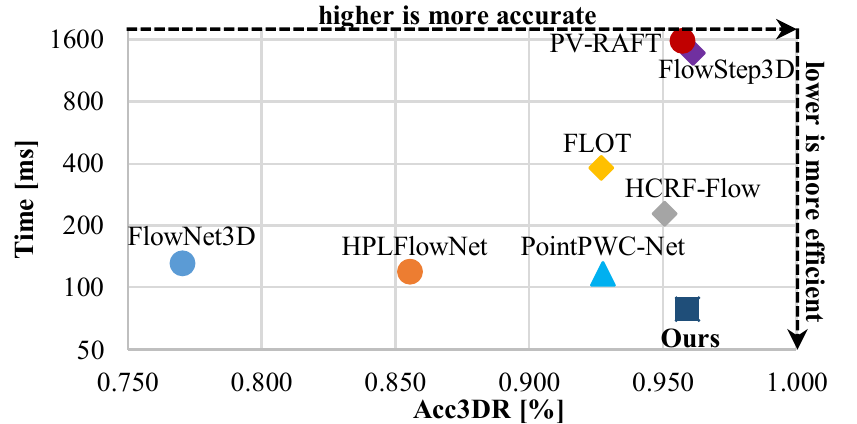}
		\caption{Our \name{} shows an accurate scene flow (Acc3DR) with less time-consuming. The accuracy is tested on the \acrfull{ft3d} \cite{mayer2016large} using 8192 points as input and the run-time is analyzed for all methods equally on a Geforce GTX 1080 Ti.}
		\label{Figure1_Teaser}
	\end{center}
	\vspace{-5.3mm}
\end{figure}


 
\begin{figure}[t]
	\begin{center}
		\includegraphics[width=1.0\linewidth]{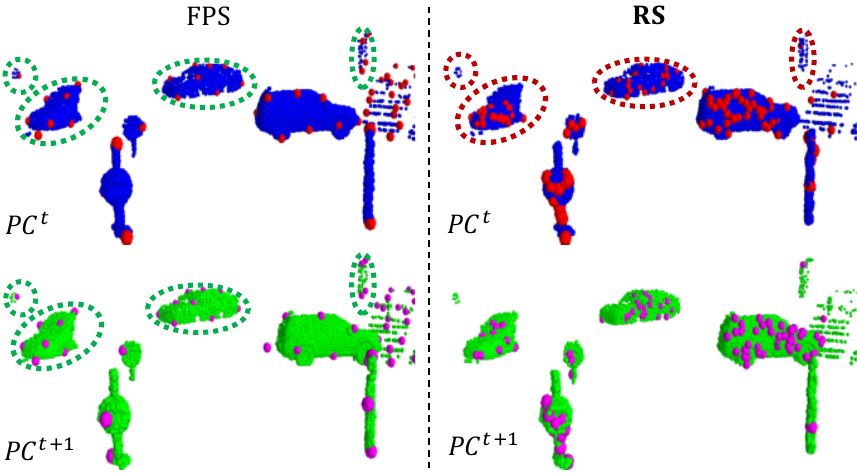}
		\caption{The challenges of \acrfull{rs} (right) compared to \acrfull{fps} (left): Both techniques sample two consecutive scenes $PC^t$ (blue) and $PC^{t+1}$ (green) into red and pink samples, respectively.}
		\label{Figure2_FPSvsRS}
	\end{center}
	\vspace{-5mm}
\end{figure}

We present in this paper our \name{} -- a hierarchical point-based learning approach that relies on \gls{rs} for scene flow estimation.
It is therefore more efficient, has a smaller memory footprint and shows comparable results at lower run-times compared to the state-of-the-art methods as shown in \cref{Figure1_Teaser}.
The use of \gls{rs} for scene flow estimation introduces big challenges and is infeasible together with existing point-based scene flow techniques \cite{wu2020pointpwc, wang2021hierarchical, kittenplon2021flowstep3d}.
This has mainly two reasons as clarified in \cref{Figure2_FPSvsRS}.
1.) \Gls{rs} will reflect the spatial distribution of the input point cloud, which is problematic if it is far from uniform.
2.) Corresponding (rigid) areas between consecutive point clouds will be sampled differently by \gls{rs}, while \gls{fps} will yield more similar patterns.

To overcome these problems, we propose a novel \PTDP{} flow embedding, which consists of three embedding layers with lateral connections (see \cref{Figure5_FlowEmbedding}) to incorporate a larger receptive field during matching.
Overall, our fully supervised architecture utilizes \gls{rs} and consists of a hierarchical feature extraction, an optimized flow embedding and scene flow predictions on multiple scales.
Our contribution is summarized as follows:
\begin{itemize}
	\item We propose \name{} -- an end-to-end scene flow estimation network that operates on dense point clouds with high accuracy.
	\item Our network uses \acrlong{rs} for a hierarchical scene flow prediction in a multi-scale fashion.
	\item We present a novel flow embedding block (called \PTDP{}) which is suitable for the combination with \acrlong{rs}.
	\item Exhaustive experiments show the strong results in terms of accuracy, generalization and run-time over the state-of-the-art methods.
\end{itemize}

\section{Related Work}


\textbf{Learning-based scene flow from point clouds:} 
Estimation of the scene flow from point clouds is a sub-field that became prominent with the availability of accurate LiDARs.
In this domain, PointFlowNet \cite{behl2019pointflownet} learns scene flow as a rigid motion coupled with object detection.
Focusing more on point-based learning with a single flow embedding, FlowNet3D \cite{liu2019flownet3d} proposes a learning-based architecture based on PointNet++ \cite{qi2017pointnet++} and MeteorNet \cite{liu2019meteornet} adds more aspects by aggregating features from spatiotemporal neighbor points.
PointPWC-Net \cite{wu2020pointpwc} is the first point-based approach that predicts scene flow hierarchically based on \cite{wu2019pointconv} without structuring or ordering them.
Despite of its high accuracy, the designed architecture is computationally expensive because of \gls{fps} with more memory consumption.
Utilizing \gls{fps}, FlowStep3D \cite{kittenplon2021flowstep3d} computes scene flow at the coarsest level and updates it iteratively through the Gated Recurrent Unit \cite{cho2014properties}.
However, this method is computationally further expensive due to the iterative update.  
Unlike the aforementioned methods, our design uses \gls{rs} instead of the expensive \gls{fps} over all its modules presenting superior efficiency and accurate results. 

Alternatively, some structure-based learning methods are employed for scene flow estimation.
In this context, Ushani \etal \cite{ushani2018feature} present a real-time method by constructing occupancy grids and HPLFlowNet \cite{gu2019hplflownet} orders the points using a permutohedral lattice.
Although their efficiency, the accuracy of such methods are limited.
Different from structure-based learning methods, our \name{} relies on point-based learning and exhibits more accurate results than the aforementioned methods at lower run-time.

Some other methods \cite{li2021self, pontes2020scene, mittal2020just} lean themselves to self-supervised category having less accuracy than our \name{}, which is designed in a fully supervised manner.


\textbf{Flow embeddings:} 
A flow embedding is a crucial part for the computation of scene flow. 
It focuses on the correlation and aggregation of corresponding features across subsequent measurements to encode the spatial displacements.
In this context, FlowNet3D \cite{liu2019flownet3d} proposes a single patch-to-point embedding block by searching 64 nearest neighbors across the consecutive point cloud at a low resolution, followed by a maximum pooling and a series of propagation and refinement blocks.
A patch-to-patch correlation is used by HPLFlowNet \cite{gu2019hplflownet} for 3D point clouds through the lattice representation. 
Recently, PointPWC-Net \cite{wu2020pointpwc} aggregated patch-to-patch features from unstructured point clouds based on a continuous weighting \cite{wu2019pointconv} which is computationally heavy. 
Utilizing the pyramid architecture as in \cite{wu2020pointpwc}, HCRF-Flow \cite{li2021hcrf} adds a high-order conditional random fields (CRFs) \cite{ristovski2013continuous} as a refinement module to explore both point-wise smoothness and region-wise rigidity. 

Utilizing \gls{fps}, HALFlow \cite{wang2021hierarchical} proposes a hierarchical attention mechanism for flow embedding.

Recently, FLOT \cite{puy20flot} built a model utilizing optimal transport based on global matching \cite{titouan2019optimal} without the use of any sampling techniques. 
Inspired by RAFT \cite{teed2020raft} to construct all-pair correlation fields, FlowStep3D \cite{kittenplon2021flowstep3d} proposes point-based and PV-RAFT \cite{wei2020pv} computes point-voxel correlation fields.  

Different from all these methods, we propose a novel and efficient \PTDP{} flow embedding block, which works reliably together with \gls{rs} without sacrificing accuracy. 

\section{Network Design} 

Our \name{} predicts scene flow from two consecutive scans of point clouds. 
These point cloud sets are ${PC^t = \{pc^t_{i}\in \mathbb{R}^3\}}^N_{i=1}$ at timestamp $t$ and ${PC^{t+1}=\{pc^{t+1}_{j}\in \mathbb{R}^3\}}^M_{j=1}$ at timestamp $t+1$, whereas ($pc^t_{i}$, $pc^{t+1}_{j}$) are the 3D Cartesian locations and ($N$, $M$) are the sizes of each set.
Our network is invariant to random permutations of the point sets.

\name{} seeks to find the similarities between point clouds to estimate the motion as scene flow vectors ${SF^t = \{sf^t_{i}\in \mathbb{R}^3 \}}^N_{i=1}$ with respect to the reference view at timestamp $t$, \ie $sf^t_{i}$ is the motion vector for $pc^t_{i}$.
The model is designed to predict the scene flow at multiple levels through hierarchical feature extraction, flow embedding, warping and scene flow estimation.
The following sections describe the components of each module in detail.     

\subsection{Feature Extraction Module} \label{feature_extraction}

\begin{figure}[t]
	\begin{center}
		\includegraphics[width=1.0\linewidth]{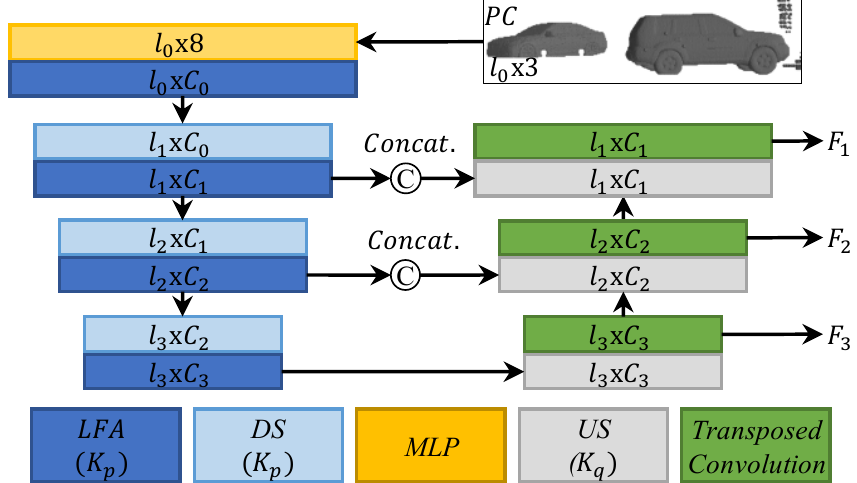}
		\caption{Our feature extraction consists two pathways: The top-down pathway constructed of \acrfull{lfa} and \acrfull{ds} with maximum pooling. The bottom-up pathway consists of \acrfull{us} and Transposed Convolution.}
		\label{Fig3_FeatureExtraction}
	\end{center}
	\vspace{-5mm}
\end{figure}

The feature extraction module consists of a feature pyramid network to extract feature sets from $PC^t$ and $PC^{t+1}$ separately.
The construction of our module involves top-down, bottom-up pathways, and lateral skip connections between them as clarified in \cref{Fig3_FeatureExtraction}.

The top-down pathway computes a hierarchy of feature sets at four scales $L=\{l\}^3_{k=0}$ from fine-to-coarse resolution, where $l_0$ is the full input resolution and the resolution of the down-sampled clouds are fixed as ${\{\{l\}}^3_{k=1}\mid l_1 = 2048, l_2 = 728, l_3 = 320\}$.
Inspired by RandLA-Net \cite{hu2020randla}, which focuses on semantic segmentation, we exploit the efficient \gls{rs} strategy combined with \gls{lfa} \cite{hu2020randla}.
\Gls{rs} has a computational complexity of $\mathcal{O}(1)$ and is therefore much more efficient compared to $\mathcal{O}(N^2)$ of \gls{fps}.
Previous works \cite{liu2019flownet3d, wu2020pointpwc, wang2021hierarchical, kittenplon2021flowstep3d} take an advantage of \gls{fps} at the cost of expensive computations.

\Gls{lfa} is employed at all scales $l_k$ except the finest one and starts by searching $K_p=17$ neighbors with \glspl{knn} and aggregates the features with two attentive pooling layers designed as in \cite{hu2020randla}.
\Gls{ds} is used to reduce the resolution from $l_k$ level to $l_{k+1}$.
We sample randomly to the defined resolution and merge the $K_p = 17$ nearest neighbors in the higher resolution for each selected sampled with maximum pooling as shown in \cref{Fig3_FeatureExtraction}.

The bottom-up pathway in our module involves ${L=\{l\}}^3_{k=1}$ layers excluding the \gls{us} to the full input resolution.
For up-scaling from level $l_{k+1}$ to $l_k$, \acrshort{knn} is used to assign the $K_q=1$ nearest neighbor for each point of the higher resolution to the lower one, followed by transposed convolution. 
To increase  the quality of the features, lateral connections are added to each level.   
This module predicts two feature sets ${F_k^t = \{f_{ki}^{t}\in \mathbb{R}^{C_k}\}}^{l_k}_{i=1}$ and ${F_k^{t+1}=\{f_{kj}^{t+1}\in \mathbb{R}^{C_k}\}}^{l_k}_{j=1}$ for $PC_k^t$ and $PC_k^{t+1}$ respectively.
Here, $C_k$ is the feature dimension fixed as ${\{\{C\}}^3_{k=1}\mid C_1 = 128, C_2 = 256, C_3 = 512\}$.
The complete feature extraction module with output channels is visualized in \cref{Fig3_FeatureExtraction}.

\subsection{Flow Embedding} \label{Flow_embedding}
A flow embedding block across two scans is the key component for scene flow estimation. 
Using \Gls{rs} requires a special flow embedding for the mentioned difficulties in \cref{introduction}.
To overcome the challenges of \Gls{rs}, we design a flow embedding block that is different from state-of-the-art. 

In this context, we establish a novel and efficient concept, called \PTDP{}, to aggregate the relation of features. This embedding block has a larger receptive field without the need to increase the number of the nearest neighbors.
To achieve this, we combine three sequential steps with lateral connections as presented in \cref{Figure5_FlowEmbedding}, and apply the entire block at each scale.

\begin{figure}[t]
	\begin{center}
	    \includegraphics[width=1.0\linewidth]{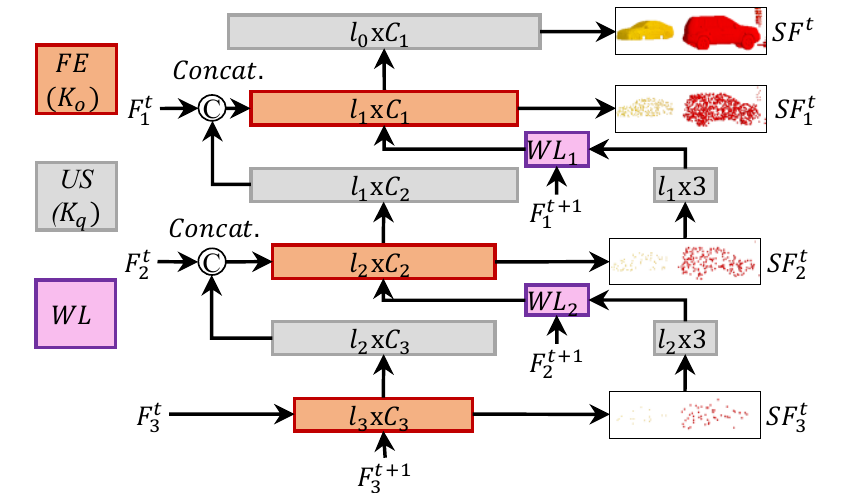}
		\caption{Multi-scale scene flow prediction with three \acrfull{fe} blocks (each consisted of three steps), two \acrfullpl{wl}, four scene flow estimators and \acrfull{us} blocks.}
		\label{Figure3_MultiScale_Prediction}
	\end{center}
    \vspace{-5mm}
\end{figure}

Starting by searching $K_o = 33$ nearest neighbors for each point $pc^t_{ki}$ within $PC_{k}^{t+1}$ at each scale $l_k$, the flow embedding consists of the following:
\begin{itemize}
	\item {$1^{st}$ Embedding} (\textit{Patch-to-Point}): It starts by grouping $K_o$ nearest features of ${F_{k}^{t+1}}$ with each point $pc^t_{ki}$. Thereafter, these grouped features will be passed into two Multi-Layer Perceptrons (MLPs) followed by maximum pooling for feature aggregation. Each MLP yields features of $C_k$ dimensions at scale $l_k$. 
	\item {$2^{nd}$ Embedding} (\textit{Point-to-Patch}): It aggregates the $K_p$ nearest features within the reference point cloud into each $pc^t_{ki}$ by computing attention scores and summation, \ie the features are weighted.
	\item {$3^{rd}$ Embedding} (\textit{Point-to-Dilated-Patch}): It repeats the previous step on the previous result with new attention scores for the $K_p$ nearest features. This embedding layer results in an increased receptive field.
\end{itemize}
Technically, we do not increase the number of the nearest neighbors for the $3^{rd}$ Embedding, but we aggregate features from a larger area by repeating the aggregation mechanism (see \cref{Figure5_FlowEmbedding}).
Overall, the three steps result in our novel \PTDP{} embedding.
This way, we are able to obtain a larger receptive field with a small number of nearest neighbors, which is computationally more efficient.
\begin{figure*}[t]
	\begin{center}
		\includegraphics[width=1.0\linewidth]{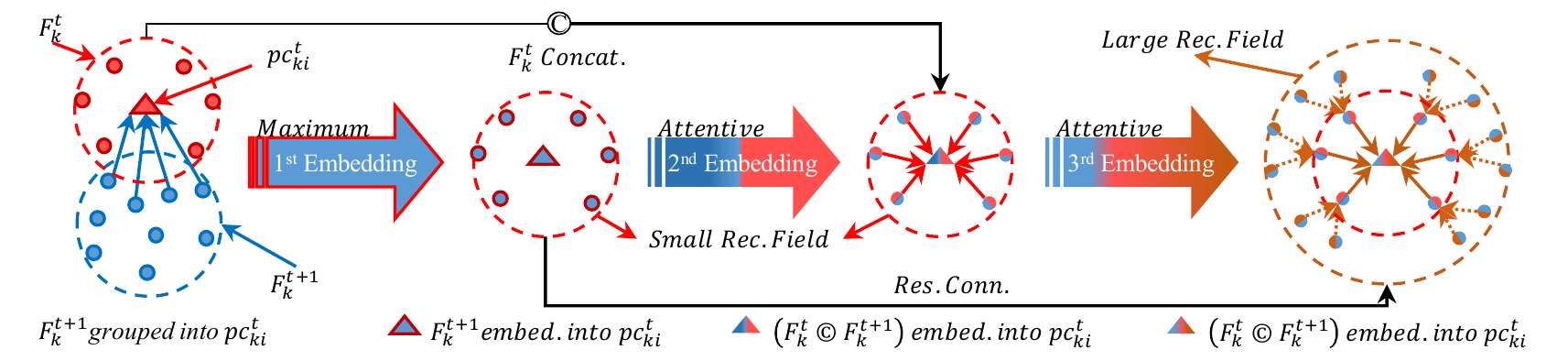}
		\caption{Our novel \acrfull{fe} block consists of three main steps: A maximum embedding across point clouds followed by two attentive embedding layers. It further uses lateral connections: A feature concatenation ($F^t_k~Concat.$) and a residual connection ($Res.~Conn.$).}
		\label{Figure5_FlowEmbedding}
	\end{center}
     \vspace{-5mm}
\end{figure*}


The attention-based aggregation technique \cite{yang2020robust, zhang2019pcan} learns attention scores for each embedded feature $f_{ki}^{t}$, followed by a \textit{softmax} to suppress the least correlated features. 
Then, the features are weighted by the attention scores and summed up. 

Additionally, we concatenate the features $F_k^t$ and add a residual connection (Res. Conn.) to increase the quality of our flow embedding (\cf \cref{Figure5_FlowEmbedding}).
This design is validated in the ablation study (\cref{Ablation}).

\subsection{Multi-Scale Scene Flow Estimation}
As mentioned, \name{} predicts scene flow at multiple scales inspired by PointPWC-Net \cite{wu2020pointpwc}, but we consider significant changes in the conjunction with \gls{rs} to make our prediction more efficient.
Our prediction of scene flow over all scales consists of two \glspl{wl}, three \glspl{fe}, three scene flow estimators and \acrfull{us} blocks as shown in \cref{Figure3_MultiScale_Prediction}.
Compared to the design of PointPWC-Net \cite{wu2020pointpwc}, we save one element from each category and we base our complete design attention mechanisms. 
Consequentially, we speed-up our model without sacrificing any accuracy as shown in our results, \cf in \cref{Table1_Comparison}.
The multi-scale estimation starts at the coarsest resolution by prediction $SF_3^t$ with a scene flow estimation module after a first \gls{fe}.
The estimation module consists of just three MLPs with 64, 32 and 3 output channels, respectively.
Thereafter, we up-sample the estimated scene flow as well as the coming features from the \gls{fe} to the next higher scale using \acrshort{knn} with $K_q = 1$.

Our \acrlong{wl} utilizes the up-sampled scene flow $SF_k^t$ at scale level $l_k$ to warp $F_k^{t+1}$ towards $F_k^{t}$. 
To this end, we add the predicted scene flow $SF_k^t$ to $PC_k^{t}$ to compute the warped ${\widetilde{PC}}_k^{t+1}$ and then we group the features $F_k^{t+1}$ into $F_k^{t}$ by using KNN search across $PC_k^{t+1}$ and ${\widetilde{PC}}_k^{t+1}$.
This warping is more simple and efficient compared to the process in PointPWC-Net \cite{wu2020pointpwc} which associates firstly the predicted scene flow to $PC_k^{t+1}$ by KNN search in order to warp $PC_k^{t+1}$ into $\widetilde{PC}_k^{t}$ and then grouping the features with another KNN search.

\subsection{Loss Function}
The model is a fully supervised at multiple scales, similar to PointPWC-Net \cite{wu2020pointpwc}.
If $SF_k^t$ is the predicted scene flow and the ground truth is $SF_{GT,k}^t$ at level $l_k$, then the objective can be written as:
\begin{equation}
\begin{aligned}
	\mathcal{L}(\theta) = \sum_{k=0}^{3} {\alpha}_k \sum_{i=1}^{l_k} {\| sf_{ki}^t - sf_{GT,ki}^t \|}_2,
\label{Equation1_TrainingLoss}
\end{aligned}	
\end{equation}
with ${\|.\|}_2$ denoting the $L_2$-norm and weights per scale of $\{\{{\alpha}_k\}^3_{k=0} \mid {\alpha}_0 = 0.02, {\alpha}_1 = 0.04, {\alpha}_2 = 0.08,  {\alpha}_3 = 0.16 \}$.

\begin{table*}[t]
	\caption{Quantitative results of our \name{} compared to the state-of-the-art methods on \acrshort{ft3d} \cite{mayer2016large} and KITTI \cite{menze2015object}. Our \name{} is trained and tested on 8192 points as other methods. The scores of the state-of-the-art methods are obtained from \cite{liu2019flownet3d, wu2020pointpwc, wang2021hierarchical, kittenplon2021flowstep3d, gu2019hplflownet, puy20flot, wei2020pv}. The best and the second best scores in all metrics are emboldened and underlined respectively. The run-time and the memory usage are compared equally on a Geforce GTX 1080 Ti. We mention the sampling strategy  of each method to facilitate the comparison.
Previous work either applies lattice-based scaling, avoids sampling, or applies \acrfull{fps}.
Our \name{} uses \acrfull{rs} and shows robust and accurate results at a low run-time.}
	\vspace{-5mm}
	\label{Table1_Comparison}
	\begin{center}
		\resizebox{1.\linewidth}{!}
		{
			\begin{tabular}{c|c|c|cccc|cc}
				\multirow{2}{*}{\textbf{Data set}}   
				& \multirow{2}{*}{\textbf{Model}} & \multirow{2}{*}{\textbf{Sampling}} 
				& \textbf{EPE3D}   & \textbf{Out3D}  & \textbf{Acc3DS}   & \textbf{Acc3DR}   & \textbf{Time} & \textbf{Memory}\\	
				&                                 &                                    
				& [m] & [\%] & [\%] & [\%] & [ms] & [GB] \Tstrut\Bstrut\\
				\hline
				\multirow{10}{*}{\rotatebox[origin=c]{90}{\textbf{\acrshort{ft3d_bold}~\cite{mayer2016large}}}}
				& FlowNet3D~\cite{liu2019flownet3d}    		 & FPS       & 0.114    & 0.602    & 0.413    & 0.771      & 132     & 10.85     \Tstrut\Bstrut\\
				& HPLFlowNet~\cite{gu2019hplflownet}   		 & Scaling   & 0.080    & 0.428    & 0.616    & 0.856      & 119     & 1.58  \Tstrut\Bstrut\\
				& PointPWC-Net~\cite{wu2020pointpwc}   		 & FPS       & 0.059    & 0.342    & 0.738    & 0.928      & \ul{117}     & 2.86  \Tstrut\Bstrut\\
				& FLOT~\cite{puy20flot}               		 & -         & 0.052    & 0.357    & 0.732    & 0.927      & 376     & 3.84  \Tstrut\Bstrut\\
				& HALFlow~\cite{wang2021hierarchical}  		 & FPS       & \ul{0.049}	& 0.308	   & 0.785	  & 0.947      & -       & -  \Tstrut\Bstrut\\
				& FlowStep3D~\cite{kittenplon2021flowstep3d} & FPS       & \textbf{0.046}	& \textbf{0.217}	   & \ul{0.816}	  & \textbf{0.961}	   & 1369    & \textbf{1.31}  \Tstrut\Bstrut\\
				& PV-RAFT~\cite{wei2020pv}	                 & -	     & \textbf{0.046}	& \ul{0.292}	   & \textbf{0.817}	  & \ul{0.957}	   & 1565    & 4.03  \Tstrut\Bstrut\\
				& \textbf{\name{} (Ours)}                              & \textbf{RS}        & 0.056	& 0.324	   & 0.792	  & 0.955	   & \textbf{77}      & \ul{1.39}  \Tstrut\Bstrut\\
				\hdashline
				\multirow{10}{*}{\rotatebox[origin=c]{90}{\textbf{KITTI~\cite{menze2015object}}}}
				& FlowNet3D~\cite{liu2019flownet3d}          & FPS       & 0.177     & 0.527    & 0.374    & 0.668      & 132   & 10.85      \Tstrut\Bstrut\\
				& HPLFlowNet~\cite{gu2019hplflownet}         & Scaling   & 0.117     & 0.410    & 0.478    & 0.778      & 119   & 1.58     \Tstrut\Bstrut\\
				& PointPWC-Net~\cite{wu2020pointpwc} 		 & FPS  	 & 0.069     & 0.265    & 0.728    & 0.888      & \ul{117}   & 2.86     \Tstrut\Bstrut\\
				& FLOT~\cite{puy20flot}              		 & -         & 0.056     & 0.242    & 0.755    & 0.908      & 376   & 3.84     \Tstrut\Bstrut\\
			    & HALFlow~\cite{wang2021hierarchical} 	     & FPS	     & 0.062	 & 0.249	& 0.765	   & 0.903	    & -     & -     \Tstrut\Bstrut\\
				& FlowStep3D~\cite{kittenplon2021flowstep3d} & FPS	     & \ul{0.055}	 & \textbf{0.149}	& 0.805	   & 0.925	    & 1369  & \textbf{1.31}     \Tstrut\Bstrut\\
				& PV-RAFT~\cite{wei2020pv}					 & -	     & 0.056	 & 0.216	& \textbf{0.823}	   & \ul{0.937}	    & 1565  & 4.03     \Tstrut\Bstrut\\
				& \textbf{\name{} (Ours)}								 & \textbf{RS}	     & \textbf{0.053}	 & \ul{0.203}	& \ul{0.818}	   & \textbf{0.938}	    & \textbf{77}    & \ul{1.39}     \Tstrut\Bstrut\\
			\end{tabular}
		}
	\end{center}
	\vspace{-4mm}

\end{table*}


\section{Experiments}
We run several experiments to evaluate the results of our \name{} for scene flow estimation. 
Firstly, we demonstrate the accuracy and the efficiency of \name{} compared to the state-of-the-art. 
Secondly, we verify our design choice with several analyses.
\begin{figure*}[t]
	\begin{tabular}{p{0.1cm}ccc}
		
		
		\rotatebox[origin=c]{90}{\textit{Scene + SF}} &
		\includegraphics[width=0.3\linewidth,valign=c]{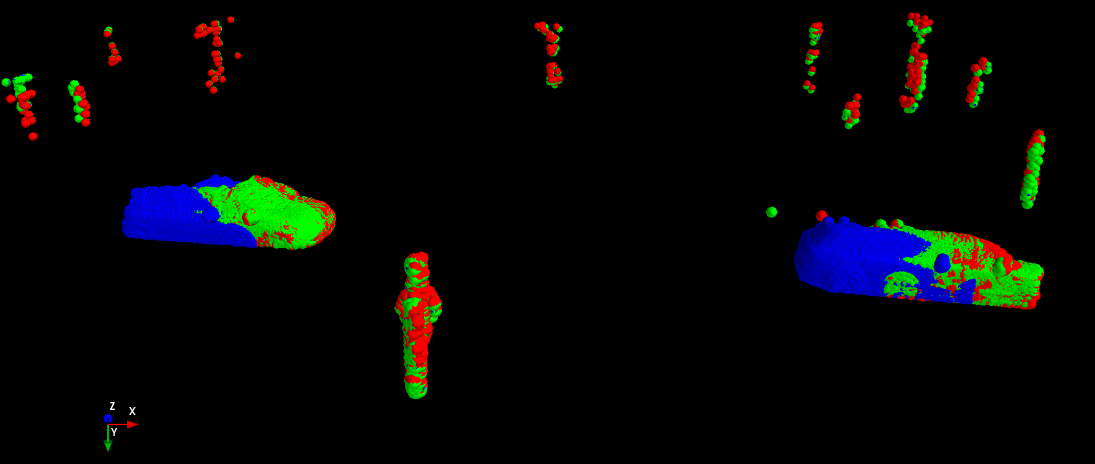}&
		\includegraphics[width=0.3\linewidth,valign=c]{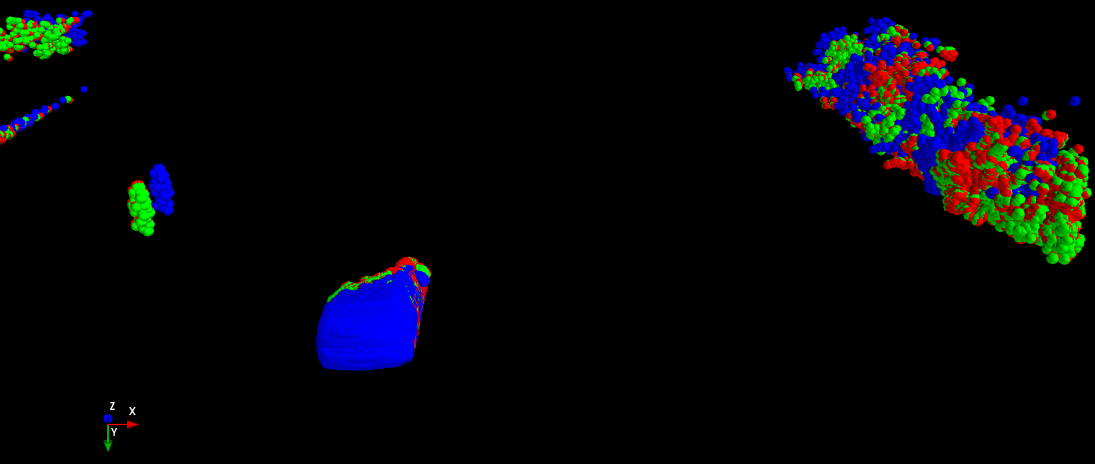}&
		\includegraphics[width=0.3\linewidth,valign=c]{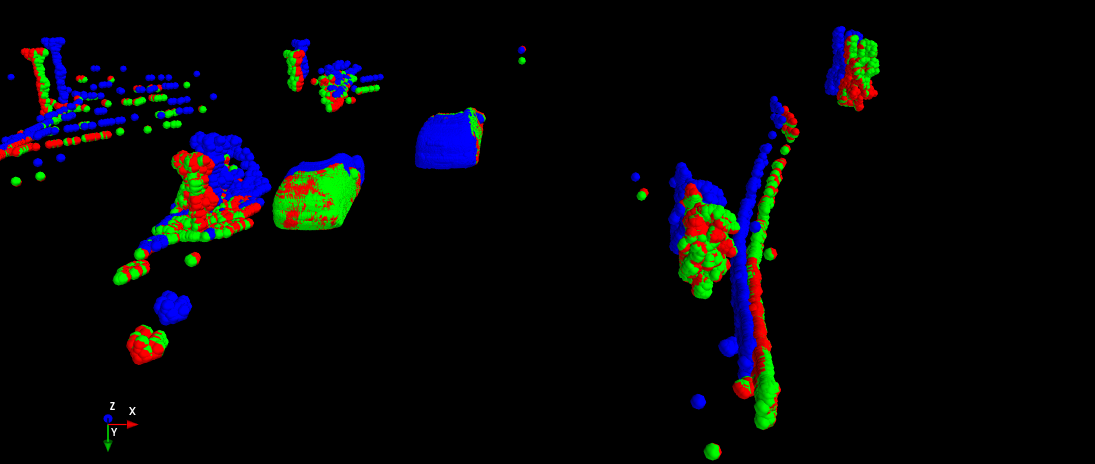}\Tstrut\Bstrut\\	
		
		\addlinespace[2pt]
		\rotatebox[origin=c]{90}{\textit{Error Map 3D}} &
		\includegraphics[width=0.3\linewidth,valign=c]{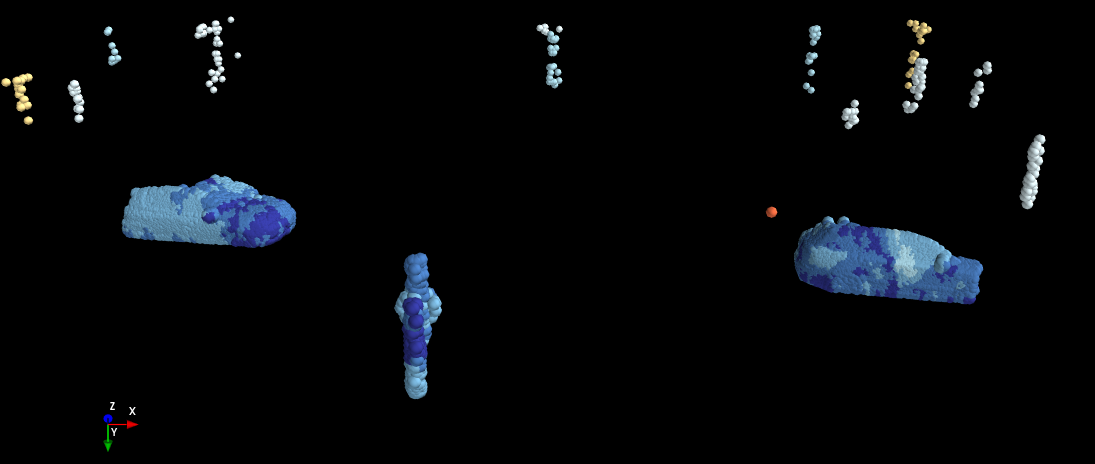}&
		\includegraphics[width=0.3\linewidth,valign=c]{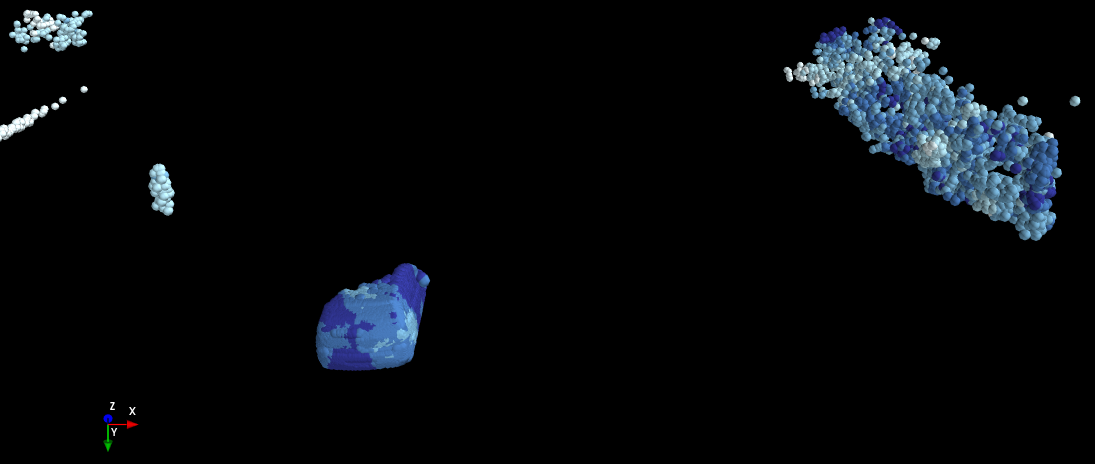}&
		\includegraphics[width=0.3\linewidth,valign=c]{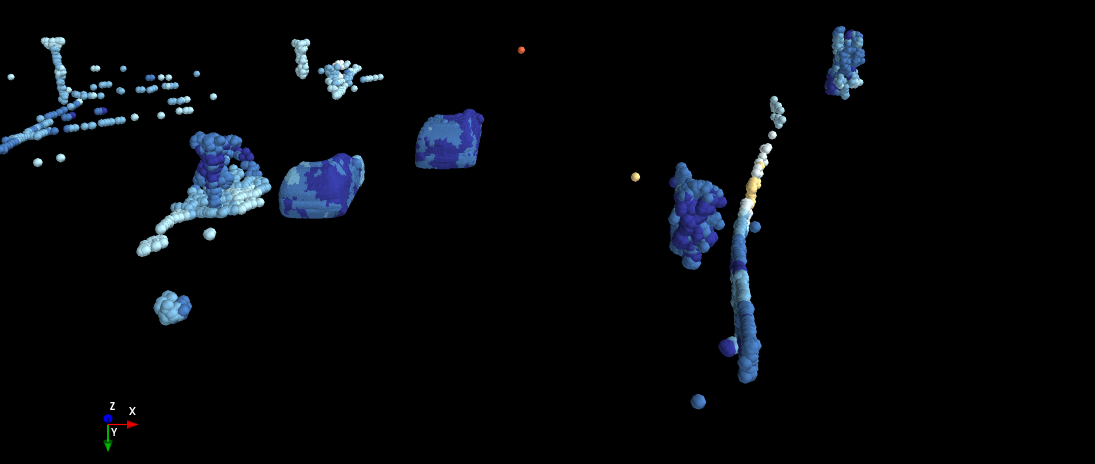}\Tstrut\Bstrut\\		
		
		\addlinespace[1pt]
		\rotatebox[origin=c]{90}{\textit{EPE3D}} &
		\multicolumn{3}{c}{\includegraphics[width=0.95\linewidth,valign=c]{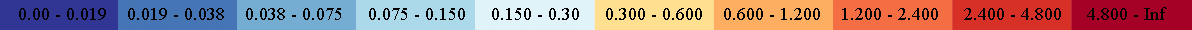}}\Tstrut\Bstrut\\
	\end{tabular}
	\caption{Three examples taken from KITTI show the impact of our \name{}. The first row of each example visualizes $PC^t$ as blue color and the predicted and ground truth scene flow after adding them to $PC^t$ in red and green color, respectively. The second row shows the end-point error in meters for each scene according to the color map shown in the last row. Our \name{} shows low errors (dark blue) across a wide range of the observed scene especially for moving objects (cars).}
	\label{Figure5_QualitativeComparison}
	\vspace{-5mm}
\end{figure*}

\subsection{Evaluation Metrics and Data Sets} \label{Data_Metrics}
For a fair comparison, we use the same evaluation metrics as in \cite{gu2019hplflownet}. Let $SF^t$ denote the predicted scene flow, and $SF_{GT}^t$ the ground truth scene flow. The evaluation metrics are averaged over all points and computed as follows:
\begin{itemize}
	\item {\textit{EPE3D [m]}}: The 3D end-point error computed in meters as ${\|SF^t-SF_{GT}^t\|}_2$. 
	\item {\textit{Acc3DS [\%]}}: The strict 3D accuracy which is the ratio of points whose EPE3D $< 0.05~m$ \textbf{or} relative error $< 5\%$. 
	\item {\textit{Acc3DR [\%]}}: The relaxed 3D accuracy which is the ratio of points whose EPE3D $< 0.1~m$ \textbf{or} relative error $< 10\%$. 
	\item {\textit{Out3D [\%]}}: The ratio of outliers whose EPE3D $> 0.3~m$ \textbf{or} relative error $> 10\%$.  
\end{itemize}
We train our~\name{} on the established data set~\gls{ft3d} \cite{mayer2016large} which consists of $19640$ labeled scene flow scenes available in the training set. 
We exclude the occluded points and the points with a depth above $35$ meters as~\cite{liu2019flownet3d, wu2020pointpwc, wang2021hierarchical, kittenplon2021flowstep3d, gu2019hplflownet, puy20flot, wei2020pv} considering most of the moving objects within the scenes.

For testing, we evaluate our model on all $3824$ frames available in the test split of~\acrshort{ft3d}. 
Since~\acrshort{ft3d} scenes are only synthetic data, we verify the generalization ability of our model to real-world scenes of the KITTI \cite{menze2015object} data set without fine-tuning.
For both data sets~\acrshort{ft3d} and KITTI, the setup of evaluation is exactly the same as in related works.

Since the existing labeled data does not provide a direct representation of point cloud information (\ie 3D Cartesian locations), we follow the established pre-processing strategy of HPLFlowNet~\cite{gu2019hplflownet}\footnote{{\url{https://github.com/laoreja/HPLFlowNet}.}} which is commonly used also in the state-of-the-art methods.

For training and evaluation with a specific resolution, the pre-processed data is randomly sub-sampled to $N$ points with random order.

\subsection{Implementation and Training} \label{Implementation}
We use the Adam optimizer with default parameters and train our model with $800$ epochs divided in two phases:
To speed up the convergence of our model, we first train $120$ epochs with a fixed set of points for each frame and apply an exponentially decaying learning rate, initialized with $0.001$, then decreased with a decaying rate of $0.7$ each $10$ epochs. 
For the next $680$ epochs, the learning rate is fixed to $0.0001$ and 8192 points are sampled randomly for each frame in each iteration.

Moreover, we add geometrical augmentation as in the related works, \ie points are randomly rotated around the X, Y and Z axes by a small angle and a random translational offset is added to increase the ability of our model to generalize without fine-tuning.

\begin{figure}[t]
	\begin{center}
		\includegraphics[width=1.0\linewidth]{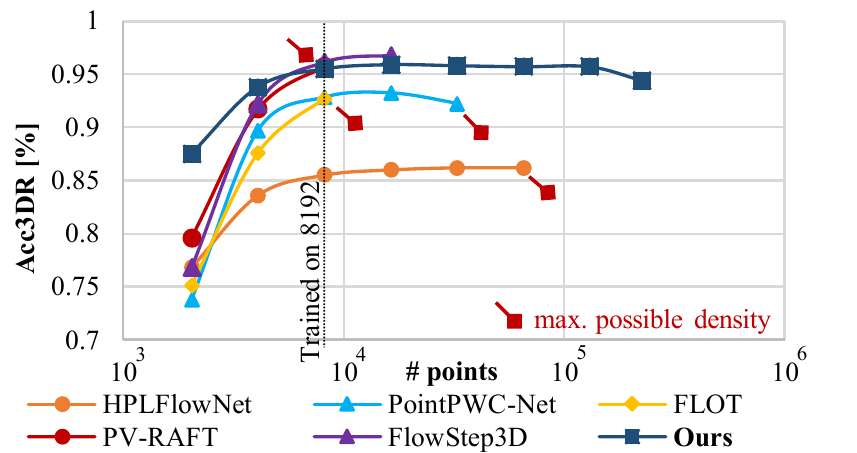}			
		\caption{Analysis of the accuracy for different numbers of points compared to state-of-the-art methods.}
		\label{Figure6_AccVsMethods}
	\end{center}
\vspace{-5mm}
\end{figure}


\subsection{Comparison to State-of-the-Art} \label{Comparison}
\textbf{Evaluation on~\acrshort{ft3d_bold}:}
In order to demonstrate the accuracy, generalization and efficiency of our model, we compare to the state-of-the-art methods in \cref{Table1_Comparison}.
Our~\name{} outperforms the methods of~\cite{liu2019flownet3d, wu2020pointpwc, gu2019hplflownet} over all evaluation metrics and shows comparable results to ~\cite{wang2021hierarchical, kittenplon2021flowstep3d, puy20flot, wei2020pv} with lowest run-time and low memory footprint. 
Compared to the concurrent methods~\cite{wang2021hierarchical, kittenplon2021flowstep3d}, which use~\gls{fps}, 
our \name{} shows comparable accuracy utilizing \gls{rs}.


\textbf{Generalization to KITTI:} 
We test the generalization ability to the KITTI data set~\cite{menze2015object} without fine-tuning. 
The reported scores in~\cref{Table1_Comparison} provide an evidence about the robustness on real-world scenes.
Our~\name{} outperforms over all the methods of~\cite{liu2019flownet3d, wu2020pointpwc, wang2021hierarchical, gu2019hplflownet, puy20flot} and presents comparable results to \cite{kittenplon2021flowstep3d, wei2020pv}.

Visually, three examples on KITTI are shown in~\cref{Figure5_QualitativeComparison} where the scene flow of a moving car and the surroundings are with low deviations compared to ground truth.

\textbf{Efficiency:}
To verify the efficiency of~\name{}, we run the official implementations of the state-of-the-art methods~\cite{liu2019flownet3d, wu2020pointpwc, kittenplon2021flowstep3d, gu2019hplflownet, puy20flot, wei2020pv} on a clean environment with a GeForce GTX 1080 Ti and measure the average inference time in milliseconds (ms) over the test set.  
As shown in \cref{Table1_Comparison}, \name{} is more efficient in terms of run-time than any other method for 8192 input points and it's near to \cite{kittenplon2021flowstep3d} in terms of memory use.
Hence, our method is~\mytilde1.5x faster than~\cite{liu2019flownet3d, wu2020pointpwc, gu2019hplflownet}, ~\mytilde4.5x faster than FLOT~\cite{puy20flot} and ~\mytilde18x faster than~\cite{kittenplon2021flowstep3d, wei2020pv} which are the main competitor to our method in terms of accuracy.

Due to the unavailability of an open source code of HALFlow \cite{wang2021hierarchical} and missing efficiency analysis in its original paper, we can not analyze the efficiency but we assume its less efficiency than us due to the use of \gls{fps}.

\begin{table}[b]
	\vspace{-4mm}
	\caption{Ablation study on various design variants for the flow embedding of our \name{}. We test all variants on 8192 points from the test split of \acrshort{ft3d} \cite{mayer2016large}.}
	\vspace{-5mm}
	\label{Table2_costvolume}
	\begin{center}
		\resizebox{1.0\linewidth}{!}
		{
			\begin{tabular}{ccccc|c}				
				$\mathbf{1^{st}}$ & $\mathbf{2^{nd}}$ & $\mathbf{F_k^t}$ & $\mathbf{Res.}$ & $\mathbf{3^{rd}}$ &  \textbf{\acrshort{ft3d_bold}~\cite{mayer2016large}} \\
				\textbf{Embed.} & \textbf{Embed.} & \textbf{Conn.} & \textbf{Concat.} & \textbf{Embed.}  & \textbf{Acc3DR} [\%]\Tstrut\Bstrut\\	
				\hline			
				\cmark & \xmark & \xmark & \xmark & \xmark & 0.792\Tstrut\Bstrut\\		
				\cmark & \cmark & \xmark & \xmark & \xmark & 0.860\Tstrut\Bstrut\\
				\cmark & \cmark & \cmark & \xmark & \xmark & 0.871\Tstrut\Bstrut\\			
				\cmark & \cmark & \cmark & \cmark & \xmark & 0.885\Tstrut\Bstrut\\
				\cmark & \cmark & \cmark & \cmark & \cmark & \textbf{0.897}\Tstrut\Bstrut\\
				\hdashline
				\multicolumn{5}{c|}{w/ full training and augmentation} & \textbf{0.955}\Tstrut\Bstrut\\
			\end{tabular}
		}
	\end{center}
\vspace{-5mm}
\end{table}

\subsection{Varying Point Densities} \label{Density}
\begin{figure}[t]
	\begin{center}
		\includegraphics[width=1.0\linewidth]{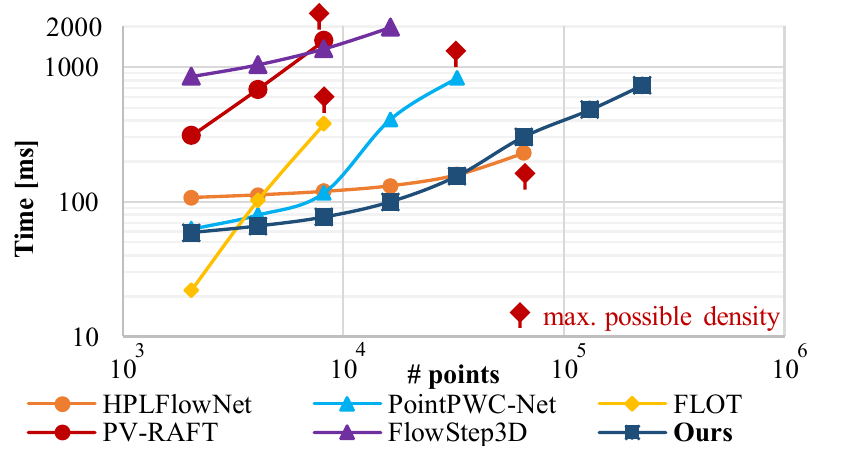}
		\caption{Run-time analysis for different numbers of points compared to state-of-the-art methods.}
		\label{Figure7_TimeVsMethods}
	\end{center}
    \vspace{-5mm}
\end{figure}

We evaluate our method against the important competitors~\cite{wu2020pointpwc, kittenplon2021flowstep3d, gu2019hplflownet, puy20flot, wei2020pv} on different point densities as shown in~\cref{Figure6_AccVsMethods}.
Acc3DR and inference time on~\gls{ft3d} are measured for a wide range of densities ${N = \{2048 * 2^i\}^6_{i=0}}$, and finally all available non-occluded points are used, which corresponds on average to \mytilde225K points (see~\cref{Figure6_AccVsMethods} and \cref{Figure7_TimeVsMethods}).
For the competing methods of FLOT~\cite{puy20flot}, PV-RAFT~\cite{wei2020pv}, PointPWC-Net~\cite{wu2020pointpwc} and HPLFlowNet~\cite{gu2019hplflownet}, the maximum possible densities are limited to $8192$, $8192$, $32768$ and $65536$, respectively, due to exceeding the memory limit of the Geforce GTX 1080 Ti for our tested range.
We have limited as well the number of points for FlowStep3D~\cite{kittenplon2021flowstep3d} to $16384$ in our test, due to the bad run-time ($>2.5$ seconds) for each frame with more dense scenes.

In contrast,~\name{} can operate on more than 250K points efficiently with high accuracy and low run-time.
In order to keep the accuracy stable for densities $> 32$K, we have increased the resolution of the down-sampled features (\cf \cref{feature_extraction}) to ${\{\{l\}}^3_{k=1}\mid l_1 = 8192, l_2 = 2048, l_3 = 512\}$ without further training or fine-tuning at all.
As a result, the accuracy remains stable over a wide range of densities (see \cref{Figure6_AccVsMethods}).
Even with this change, our~\name{} is more efficient and faster than previous works for increased input densities (see \cref{Figure7_TimeVsMethods}).
The design of~\name{} allows to operate on a much higher maximum density compared to other methods, due to their memory footprint and the time consumption.
However, the run-time of our~\name{} still increases super-linear with growing input density due to the~\acrshort{knn} search. 


\subsection{Ablation Study} \label{Ablation}
Finally, we verify our design choices for the~\gls{fe}~on~\gls{ft3d}~\cite{mayer2016large} by removing components of the~\gls{fe} and compare the variants in~\cref{Table2_costvolume}.
The models in this comparison are trained only for the first phase as explained in~\cref{Implementation} and without augmentation.
Each part clearly adds a contribution to the overall accuracy when using~\gls{rs}. 
The $1^{st}$ Embedding with maximum pooling, as basically used in FlowNet3D~\cite{liu2019flownet3d}, is not able to resolve the challenges of the \acrshort{rs} strategy for scene flow estimation.
The complete design -- three embedding layers with lateral connections -- leads to the best results.



\section{Conclusion}
In this paper, we have proposed \name{} -- an efficient and fully supervised network for multi-scale scene flow estimation in large-scale point clouds. 
Utilizing \acrfull{rs} during feature extraction, we could boost the run-time and memory footprint to allow for an efficient processing of point clouds at an unmatched maximum density.
The novel \acrlong{fe} block (called \PTDP{}), resolves the prominent challenges when using \acrshort{rs} for scene flow estimation.
Consequentially, \name{} reaches state-of-the-art accuracy on \acrshort{ft3d} and generalizes well over a wide range of input densities as well as to the real-world scenes of KITTI.

\section*{Acknowledgment}
This work was partially funded by the BMW Group and partially by the Federal Ministry of Education and Research Germany under the project DECODE (01IW21001).

\bibliographystyle{IEEEtran}
\bibliography{IEEEabrv, RMS-FlowNet}

\begin{thebibliography}{10}
\providecommand{\url}[1]{#1}
\csname url@rmstyle\endcsname
\providecommand{\newblock}{\relax}
\providecommand{\bibinfo}[2]{#2}
\providecommand\BIBentrySTDinterwordspacing{\spaceskip=0pt\relax}
\providecommand\BIBentryALTinterwordstretchfactor{4}
\providecommand\BIBentryALTinterwordspacing{\spaceskip=\fontdimen2\font plus
\BIBentryALTinterwordstretchfactor\fontdimen3\font minus
  \fontdimen4\font\relax}
\providecommand\BIBforeignlanguage[2]{{%
\expandafter\ifx\csname l@#1\endcsname\relax
\typeout{** WARNING: IEEEtran.bst: No hyphenation pattern has been}%
\typeout{** loaded for the language `#1'. Using the pattern for}%
\typeout{** the default language instead.}%
\else
\language=\csname l@#1\endcsname
\fi
#2}}

\bibitem{brickwedde2019mono}
F.~Brickwedde, S.~Abraham, and R.~Mester, ``{Mono-SF: Multi-View Geometry Meets
  Single-View Depth for Monocular Scene Flow Estimation of Dynamic Traffic
  Scenes},'' in \emph{IEEE/CVF International Conference on Computer Vision
  (CVPR)}, 2019.

\bibitem{hur2020self}
J.~Hur and S.~Roth, ``{Self-Supervised Monocular Scene Flow Estimation},'' in
  \emph{Proceedings of the IEEE/CVF Conference on Computer Vision and Pattern
  Recognition}, 2020.

\bibitem{aleotti2020learning}
F.~Aleotti, M.~Poggi, F.~Tosi, and S.~Mattoccia, ``{Learning End-To-End Scene
  Flow by Distilling Single Tasks Knowledge},'' in \emph{Conference on
  Artificial Intelligence (AAAI)}, 2020.

\bibitem{ilg2018occlusions}
E.~Ilg, T.~Saikia, M.~Keuper, and T.~Brox, ``{Occlusions, Motion and Depth
  Boundaries with a Generic Network for Disparity, Optical Flow or Scene Flow
  Estimation},'' in \emph{Proceedings of the European Conference on Computer
  Vision (ECCV)}, 2018.

\bibitem{jiang2019sense}
H.~Jiang, D.~Sun, V.~Jampani, Z.~Lv, E.~Learned-Miller, and J.~Kautz, ``{SENSE:
  a Shared Encoder Network for Scene-flow Estimation},'' in \emph{IEEE/CVF
  International Conference on Computer Vision (ICCV)}, 2019.

\bibitem{saxena2019pwoc}
R.~Saxena, R.~Schuster, O.~Wasenm{\"u}ller, and D.~Stricker, ``{PWOC-3D: Deep
  Occlusion-Aware End-to-End Scene Flow Estimation},'' \emph{IEEE International
  Conference on Intelligent Vehicles Symposium (IV)}, 2019.

\bibitem{schuster2021deep}
R.~Schuster, C.~Unger, and D.~Stricker, ``{A Deep Temporal Fusion Framework for
  Scene Flow Using a Learnable Motion Model and Occlusions},'' in
  \emph{IEEE/CVF Winter Conference on Applications of Computer Vision (WACV)},
  2021.

\bibitem{schuster2017sceneflowfields}
R.~Schuster, O.~Wasenm{\"u}ller, G.~Kuschk, C.~Bailer, and D.~Stricker,
  ``{SceneFlowFields: Dense Interpolation of Sparse Scene Flow
  Correspondences},'' in \emph{IEEE Winter Conference on Applications of
  Computer Vision (WACV)}, 2018.

\bibitem{yang2020upgrading}
G.~Yang and D.~Ramanan, ``{Upgrading Optical Flow to 3D Scene Flow through
  Optical Expansion},'' in \emph{IEEE/CVF Conference on Computer Vision and
  Pattern Recognition (CVPR)}, 2020.

\bibitem{liu2019flownet3d}
X.~Liu, C.~R. Qi, and L.~J. Guibas, ``{FlowNet3D: Learning Scene Flow in 3D
  Point Clouds},'' in \emph{IEEE/CVF Conference on Computer Vision and Pattern
  Recognition (CVPR)}, 2019.

\bibitem{wu2020pointpwc}
W.~Wu, Z.~Y. Wang, Z.~Li, W.~Liu, and L.~Fuxin, ``{PointPWC-Net: Cost Volume on
  Point Clouds for (Self-) Supervised Scene Flow Estimation},'' in
  \emph{European Conference on Computer Vision (ECCV)}, 2020.

\bibitem{wang2021hierarchical}
G.~Wang, X.~Wu, Z.~Liu, and H.~Wang, ``{Hierarchical Attention Learning of
  Scene Flow in 3D Point Clouds},'' \emph{IEEE Transactions on Image Processing
  (TIP)}, 2021.

\bibitem{kittenplon2021flowstep3d}
Y.~Kittenplon, Y.~C. Eldar, and D.~Raviv, ``{FlowStep3D: Model Unrolling for
  Self-Supervised Scene Flow Estimation},'' in \emph{IEEE/CVF Conference on
  Computer Vision and Pattern Recognition (CVPR)}, 2021.

\bibitem{li2018pointcnn}
Y.~Li, R.~Bu, M.~Sun, W.~Wu, X.~Di, and B.~Chen, ``{PointCNN: Convolution On
  $\chi$-Transformed Points},'' in \emph{Proceedings of the 32nd International
  Conference on Neural Information Processing Systems (NIPS)}, 2018.

\bibitem{qi2017pointnet++}
C.~R. Qi, L.~Yi, H.~Su, and L.~J. Guibas, ``{PointNet++: Deep hierarchical
  feature learning on point sets in a metric space},'' in \emph{Advances in
  neural information processing systems (NIPS)}, 2017.

\bibitem{wu2019pointconv}
W.~Wu, Z.~Qi, and L.~Fuxin, ``{PointConv: Deep Convolutional Networks on 3D
  Point Clouds},'' in \emph{IEEE/CVF Conference on Computer Vision and Pattern
  Recognition (CVPR)}, 2019.

\bibitem{zhao2019pointweb}
H.~Zhao, L.~Jiang, C.-W. Fu, and J.~Jia, ``{PointWeb: Enhancing Local
  Neighborhood Features for Point Cloud Processing},'' in \emph{IEEE/CVF
  Conference on Computer Vision and Pattern Recognition (CVPR)}, 2019.

\bibitem{mayer2016large}
N.~Mayer, E.~Ilg, P.~Hausser, P.~Fischer, D.~Cremers, A.~Dosovitskiy, and
  T.~Brox, ``A large dataset to train convolutional networks for disparity,
  optical flow, and scene flow estimation,'' in \emph{IEEE International
  Conference on Computer Vision and Pattern Recognition (CVPR)}, 2016.

\bibitem{behl2019pointflownet}
A.~Behl, D.~Paschalidou, S.~Donn{\'e}, and A.~Geiger, ``{PointFlowNet: Learning
  Representations for Rigid Motion Estimation from Point Clouds},'' in
  \emph{IEEE/CVF Conference on Computer Vision and Pattern Recognition (CVPR)},
  2019.

\bibitem{liu2019meteornet}
X.~Liu, M.~Yan, and J.~Bohg, ``{MeteorNet: Deep Learning on Dynamic 3D Point
  Cloud Sequences},'' in \emph{IEEE/CVF International Conference on Computer
  Vision (ICCV)}, 2019.

\bibitem{cho2014properties}
K.~Cho, B.~Van~Merri{\"e}nboer, D.~Bahdanau, and Y.~Bengio, ``{On the
  properties of neural machine translation: Encoder-decoder approaches},''
  \emph{arXiv preprint arXiv:1409.1259}, 2014.

\bibitem{ushani2018feature}
A.~K. Ushani and R.~M. Eustice, ``{Feature Learning for Scene Flow Estimation
  from LIDAR},'' in \emph{Conference on Robot Learning (CoRL)}, 2018.

\bibitem{gu2019hplflownet}
X.~Gu, Y.~Wang, C.~Wu, Y.~J. Lee, and P.~Wang, ``{HPLFlowNet: Hierarchical
  Permutohedral Lattice FlowNet for Scene Flow Estimation on Large-scale Point
  Clouds},'' in \emph{IEEE/CVF Conference on Computer Vision and Pattern
  Recognition (CVPR)}, 2019.

\bibitem{li2021self}
R.~Li, G.~Lin, and L.~Xie, ``{Self-Point-Flow: Self-Supervised Scene Flow
  Estimation from Point Clouds with Optimal Transport and Random Walk},'' in
  \emph{IEEE/CVF Conference on Computer Vision and Pattern Recognition (CVPR)},
  2021.

\bibitem{pontes2020scene}
J.~K. Pontes, J.~Hays, and S.~Lucey, ``Scene flow from point clouds with or
  without learning,'' in \emph{International Conference on 3D Vision (3DV)},
  2020.

\bibitem{mittal2020just}
H.~Mittal, B.~Okorn, and D.~Held, ``{Just Go with the Flow: Self-Supervised
  Scene Flow Estimation},'' in \emph{IEEE/CVF Conference on Computer Vision and
  Pattern Recognition (CVPR)n}, 2020.

\bibitem{li2021hcrf}
R.~Li, G.~Lin, T.~He, F.~Liu, and C.~Shen, ``{HCRF-Flow: Scene Flow from Point
  Clouds with Continuous High-order CRFs and Position-aware Flow Embedding},''
  in \emph{IEEE/CVF Conference on Computer Vision and Pattern Recognition
  (CVPR)}, 2021.

\bibitem{ristovski2013continuous}
K.~Ristovski, V.~Radosavljevic, S.~Vucetic, and Z.~Obradovic, ``{Continuous
  Conditional Random Fields for Efficient Regression in Large Fully Connected
  Graphs},'' in \emph{Conference on Artificial Intelligence (AAAI)}, 2013.

\bibitem{puy20flot}
G.~Puy, A.~Boulch, and R.~Marlet, ``{FLOT}: {S}cene {F}low on {P}oint {C}louds
  {G}uided by {O}ptimal {T}ransport,'' in \emph{European Conference on Computer
  Vision (ECCV)}, 2020.

\bibitem{titouan2019optimal}
V.~Titouan, N.~Courty, R.~Tavenard, and R.~Flamary, ``{Optimal Transport for
  structured data with application on graphs},'' in \emph{International
  Conference on Machine Learning}, 2019.

\bibitem{teed2020raft}
Z.~Teed and J.~Deng, ``{RAFT: Recurrent all-pairs field transforms for optical
  flow},'' in \emph{European conference on computer vision (ECCV)}, 2020.

\bibitem{wei2020pv}
Y.~Wei, Z.~Wang, Y.~Rao, J.~Lu, and J.~Zhou, ``{PV-RAFT: Point-Voxel
  Correlation Fields for Scene Flow Estimation of Point Clouds},'' in
  \emph{IEEE/CVF Conference on Computer Vision and Pattern Recognition (CVPR)},
  2021.

\bibitem{hu2020randla}
Q.~Hu, B.~Yang, L.~Xie, S.~Rosa, Y.~Guo, Z.~Wang, N.~Trigoni, and A.~Markham,
  ``{RandLA-Net: Efficient Semantic Segmentation of Large-Scale Point
  Clouds},'' in \emph{IEEE/CVF Conference on Computer Vision and Pattern
  Recognition (CVPR)}, 2020.

\bibitem{yang2020robust}
B.~Yang, S.~Wang, A.~Markham, and N.~Trigoni, ``{Robust Attentional Aggregation
  of Deep Feature Sets for Multi-view 3D Reconstruction},'' \emph{International
  Journal of Computer Vision (IJCV)}, 2020.

\bibitem{zhang2019pcan}
W.~Zhang and C.~Xiao, ``{PCAN: 3D Attention Map Learning Using Contextual
  Information for Point Cloud Based Retrieval},'' in \emph{IEEE/CVF Conference
  on Computer Vision and Pattern Recognition (CVPR)}, 2019.

\bibitem{menze2015object}
M.~Menze and A.~Geiger, ``Object scene flow for autonomous vehicles,'' in
  \emph{IEEE International Conference on Computer Vision and Pattern
  Recognition (CVPR)}, 2015.

\end{thebibliography}

\end{document}